\documentclass[letterpaper, 10 pt, conference]{ieeeconf}  

\IEEEoverridecommandlockouts                              

\usepackage{times}
\usepackage{multicol}
\usepackage[bookmarks=true]{hyperref}
\usepackage{amsmath, amssymb}
\usepackage{bbm}
\usepackage{bm}
\usepackage{graphicx}
\usepackage[linesnumbered,ruled]{algorithm2e}
\usepackage{multirow}
\usepackage{booktabs}
\usepackage{dblfloatfix}
\usepackage[dvipsnames]{xcolor}
\usepackage[T1]{fontenc}
\usepackage{dsfont}
\usepackage{color, colortbl}
\usepackage{comment}

\usepackage{hyperref}
\hypersetup{
  colorlinks,
  citecolor=Violet,
  linkcolor=OrangeRed,
  urlcolor=Blue}

\newcommand{\fig}[1]{Fig.~\ref{#1}} 
\newcommand{\etal}{\textit{et al}.}

\definecolor{Gray}{gray}{0.9}

\newcommand{\BB}{\mathcal{B}}
\newcommand{\SC}{\mathcal{S}}

\title{\LARGE \bf
BayesOD: A Bayesian Approach for Uncertainty Estimation in Deep Object Detectors}

\author{Ali Harakeh$^{1}$, Michael Smart$^{2}$, and Steven L. Waslander$^{1}$
\thanks{$^{1}$Ali Harakeh and Steven L. Waslander are with The Institute For Aerospace Studies (UTIAS), University of Toronto, Toronto, Canada, {\tt\small ali.harakeh@utoronto.ca, steven.w@utias.utoronto.ca}}%
\thanks{$^{2}$ Michael Smart is with the Department of Mechanical and Mechatronics Engineering, University of Waterloo, Waterloo, Canada, {\tt\small michael.smart@uwaterloo.ca}}%
}

\begin{document}

\maketitle
\thispagestyle{empty}
\pagestyle{empty}

\begin{abstract}
When incorporating deep neural networks into robotic systems, a major challenge is the lack of uncertainty measures associated with their output predictions. Methods for uncertainty estimation in the output of deep object detectors (DNNs) have been proposed in recent works, but have had limited success due to 1) information loss at the detectors non-maximum suppression (NMS) stage, and 2) failure to take into account the multitask, many-to-one nature of anchor-based object detection. To that end, we introduce BayesOD, an uncertainty estimation approach that reformulates the standard object detector inference and Non-Maximum suppression components from a Bayesian perspective. Experiments performed on four common object detection datasets show that BayesOD provides uncertainty estimates that are better correlated with the accuracy of detections, manifesting as a significant reduction of 9.77\%-13.13\% on the minimum Gaussian uncertainty error metric and a reduction of  1.63\%-5.23\% on the minimum Categorical uncertainty error metric. Code will be released at {\url{https://github.com/asharakeh/bayes-od-rc}}. 
\end{abstract}

\section{Introduction}
\label{sec:intro}
\begin{figure}[t]
    \centering
    \includegraphics[width=\columnwidth]{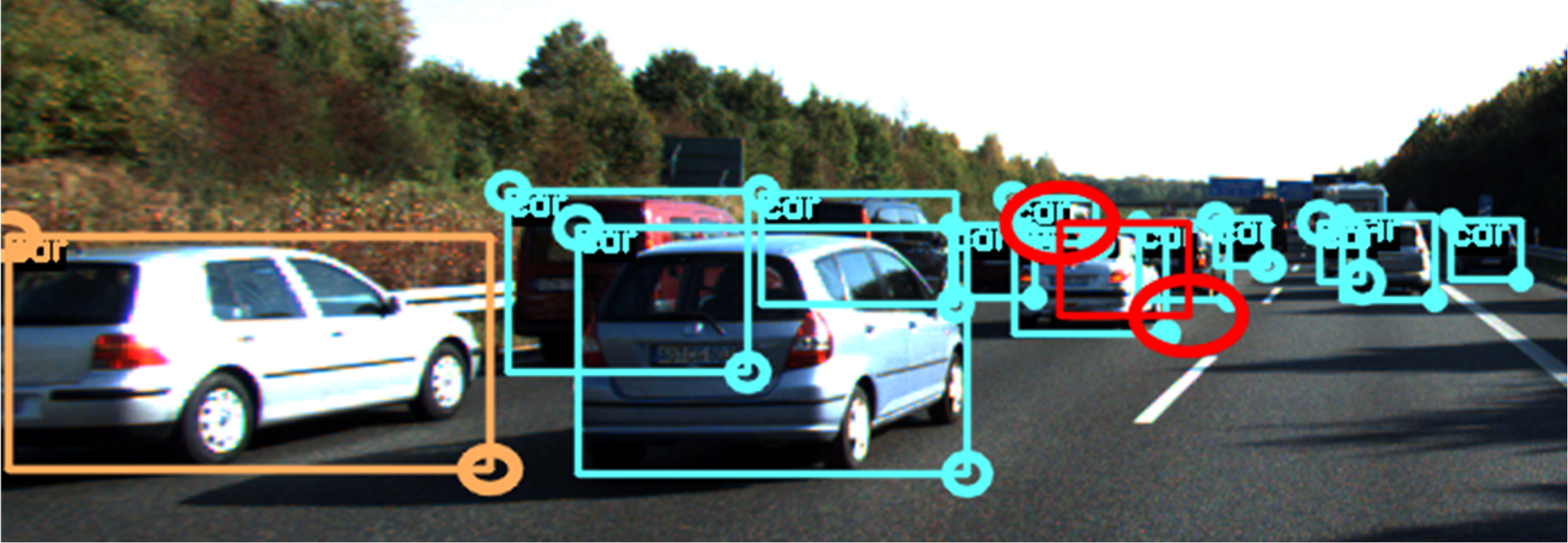}
    \caption{The output from BayesOD, demonstrated on a test image frame from the KITTI Dataset~\cite{Kitti_Geiger_2012_CVPR}. Three levels of trust (\textbf{teal}: highly reliable, \textbf{orange}: slightly reliable and \textbf{red}: unreliable) are determined based on thresholds of the Gaussian entropy provided by BayesOD. All bounding boxes are shown with the 95\% confidence ellipse of their top-left and bottom-right corners.}
    \vspace{-1.5em}
    \label{fig:intro}
\end{figure}

Due to their high level of performance, deep object detectors have become standard components of perception stacks for safety critical tasks such as autonomous driving~\cite{AVOD_Ku_2018_IROS, VoxelNet_Zhou_2018_CVPR, MV3D_Chen_2017_CVPR} and automated surveillance~\cite{Suspecious_Pandit_ICICT_2016}. Therefore, the quantification of how trustworthy these detectors are for subsequent modules, especially in safety critical systems, is of utmost importance. To encode the level of confidence in an estimate, a meaningful and consistent measure of uncertainty should be provided for every detection instance (see \fig{fig:intro}).

Two important goals must be met to create a meaningful uncertainty measure. First, the robotic system should be capable of using the uncertainty measure to fuse an object detector's output with prior information from different sources~\cite{Deep_Robotics_Saunderhauf_2018_IJRR} to connect sequences of detections over time and increase detection and tracking performance as a result. Second and most importantly, the robotic system should be able to use its own estimates of detection uncertainty to reliably identify incorrect detections, including those resulting from \textit{out of distribution instances}, where object categories, scenarios, textures, or environmental conditions have not been seen during the training phase~\cite{Deep_Robotics_Saunderhauf_2018_IJRR}.  

Two sources of uncertainty can be identified in any machine learning model. \textit{Epistemic} or model uncertainty is the uncertainty in the model's parameters, usually as a result of the confusion about which model generated the training data, and can be explained away given enough representative training data points~\cite{MC_Dropout_Gal_2016_ICML}. \textit{Aleatoric} or observation uncertainty results from the stochastic nature of the observed input, and persists in network output despite expanded training on additional data~\cite{What_Uncertainty_Kendal_2017_NIPS}.

Methods to estimate both uncertainty types in DNNs have been recently proposed by Kendal \etal ~\cite{What_Uncertainty_Kendal_2017_NIPS}, with applications to pixel-wise perception tasks. Recent methods ~\cite{Dropout_Miller_2018_ICRA, Evaluating_Merging_Miller_2018_Arxiv,LaserNet_Mayer_CVPR_2019, Bounding_Box_He_CVPR_2019, Towards_Feng_ITSC_2018,Leveraging_Feng_2018_ITSC,Uncertainty_Le_2018_ITSC} extended Kendal's work ~\cite{What_Uncertainty_Kendal_2017_NIPS} to object detection, but fail to consider the multi-task, many-to-one nature of the object detection task. To that end, we introduce BayesOD, a framework designed to estimate the uncertainty in both bounding box and category of detected object instances. This paper offers the following contributions:
\begin{itemize}
    \item We provide a Bayesian treatment for every step of the neural network inference procedure, allowing the incorporation of anchor-level and object-level priors in closed form.
    \item We replace standard non-maximum suppression (NMS) with Bayesian inference, allowing the detector to retain all predicted information for \textbf{both the bounding box and the category} of a detected object instance.
    \item We perform comprehensive experiments to quantify the quality of the estimated uncertainty on four commonly used 2D object detection datasets, COCO, Pascal VOC, Berkeley Deep Drive and Kitti. We show that BayesOD provides a significant reduction of $9.77\% - 13.13\%$ on the minimum Gaussian uncertainty error metric, a reduction of  $1.63\%-5.23\%$ on the minimum Categorical uncertainty error metric, and an increase of $0.07 \% - 3.00\% $ on the probabilistic detection quality over the next best method from current state of the art.
\end{itemize}

\begin{figure*}[t]
    \centering
    \includegraphics[width=\textwidth,trim=15 15 15 15,clip]{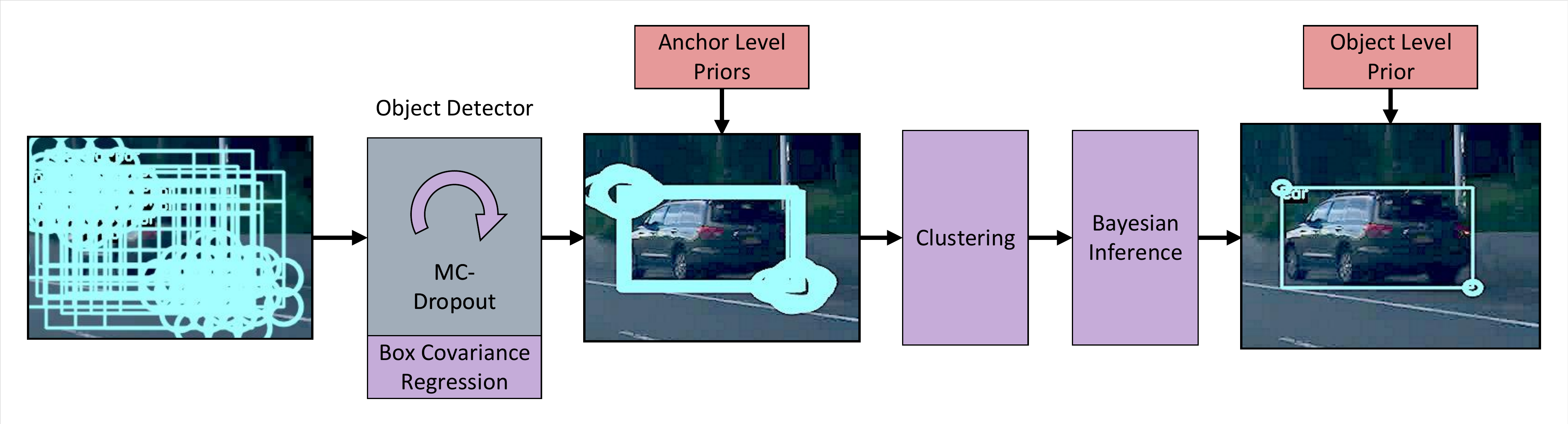}
    \vspace{-1.5em}
    \caption{The different stages of estimation employed in BayesOD, demonstrated on a test image frame from the BDD Dataset~\cite{BDD_100K_Yu_2018_Arxiv}. The additions by BayesOD to a standard object detector (\textbf{grey}) are shown in \textbf{purple}. Prior information is shown in \textbf{red}. \textbf{Left:} Prior bounding boxes. \textbf{Middle:} Object detector results after processing the prior boxes and incorporating anchor-level non-informative priors. \textbf{Right:} Final detection results after clustering and Bayesian Inference. Box corner covariance is visualized as in \fig{fig:intro}}
    \vspace{-1.5em}
    \label{fig:bayesian_framework}
\end{figure*}

\section{Related Work}
\label{sec:lit_review}

\subsection{Deep Neural Networks For Object Detection}
The object detection problem requires the estimation of both the category to which an object belongs, and its spatial location and extent, often expressed as the tightest fitting bounding box. The majority of state of the art object detectors in 2D~\cite{Speed_Accuracy_Huang_2017_CVPR} or in 3D~\cite{AVOD_Ku_2018_IROS, VoxelNet_Zhou_2018_CVPR, MV3D_Chen_2017_CVPR} follow a standard algorithm, which maps a scene representation to object instances. Since the number of object instances in the scene is usually unknown a priori, the procedure begins with a densely sampled grid of prior object bounding boxes, referred to as \textit{anchors}~\cite{Faster_R_CNN_Shaoqing_2015_NIPS, Focal_Loss_Lin_2017_ICCV}, where the object detector provides a category and a bounding box estimate for each anchor element. Since multiple anchors can be mapped to a single bounding box in space, redundant outputs are eliminated through Non-Maximum Suppression. BayesOD builds on the RetinaNet 2D object detector~\cite{Focal_Loss_Lin_2017_ICCV}.

\subsection{Uncertainty Estimation In Deep Object Detectors}
To account for epistemic uncertainty, Bayesian Neural Networks~\cite{Probable_Makey_1995_CNS} usually apply a prior distribution over their parameters $\bm \theta$ to compute a posterior distribution $p(\bm\theta|\mathcal{D})$ over the set of all possible parameters given the training dataset $\mathcal{D}$. A marginal distribution can then be computed for any prediction as:
\begin{equation}
\label{eq:parameter_marginal}
p(\bm{\hat y}_i| \textbf{x}_i,\mathcal{D}) = \int\limits_\theta p(\bm{\hat y}_i| \textbf{x}_i,\mathcal{D}, \bm{\theta})p(\bm\theta|\mathcal{D})d\bm\theta, \end{equation}
where $\textbf{x}_i$ is the input, and $\bm{\hat y}_i$ is the output of the neural network. Unfortunately, the calculation of the integral in Eq.~\eqref{eq:parameter_marginal} is usually intractable due to the non-linear activation function between consecutive layers~\cite{Evidential_Sensoy_2018_NIPS}. Tractable approximations can be derived through Monte-Carlo integration by using ensemble methods~\cite{Lakshminarayanan_Ensembles_NIPS_2017} or Monte Carlo (MC) Dropout~\cite{MC_Dropout_Gal_2016_ICML}.

To estimate the epistemic uncertainty in the output of deep object detectors, Miller  \etal~\cite{Dropout_Miller_2018_ICRA} directly applies MC Dropout, treating the deep object detector as a \textbf{black box}. Uncertainty is then estimated as sample statistics from spatially correlated detector outputs. Subsequent work~\cite{Evaluating_Merging_Miller_2018_Arxiv} studied the effect of various correlation and merging algorithms on the quality of the estimated uncertainty measures from the black box method in~\cite{Dropout_Miller_2018_ICRA}. The black box method is shown to provide weakly correlated estimates for bounding box uncertainty, mainly because it observes the output bounding box after NMS, where most of the information from redundant predictions has already been removed.

Kendall \etal~\cite{What_Uncertainty_Kendal_2017_NIPS} provides one of the first works to address the estimation of aleatoric uncertainty for computer vision tasks. For regression tasks, a log likelihood loss is used to estimate heteroscedastic aleatoric uncertainty, written for every regression target as:
\begin{equation}
\label{eq:kendal_variance_regression}
L_{reg}(\textbf{x}, \bm\theta) =  \frac{1}{2\sigma(\textbf{x}, \bm\theta)^2} ||\textbf{y} - f(\textbf{x}, \bm\theta)||^2_2 + \frac{1}{2}\log \sigma(\textbf{x}, \bm\theta)^2,
\end{equation}
where $\textbf{x}$ is the input to, and $f(\textbf{x}, \bm\theta)$ is the output from the neural network. Furthermore $\textbf{y}$ is the ground truth regression target, $||.||_2$ is the $L_2$ norm, $\bm\theta$ are the neural network parameters, and $\sigma(\textbf{x}, \bm\theta)$ is the \textbf{estimated} output variance.

Le \etal~\cite{Uncertainty_Le_2018_ITSC} directly apply the formulation in Eq.~\eqref{eq:kendal_variance_regression} to estimate the diagonal elements of the covariance matrix of the bounding box output from object detectors. Such methods are referred to as \textbf{sampling free} and require only a single run of the deep object detector to estimate uncertainty. The estimated variance in Eq.~\eqref{eq:kendal_variance_regression} has also been used in~\cite{LaserNet_Mayer_CVPR_2019, Bounding_Box_He_CVPR_2019, Leveraging_Feng_2018_ITSC} to increase average precision, by incorporating it in the non-maximum suppression stage, while disregarding the quality of the output uncertainty. The proposed sampling free methods assume a diagonal covariance matrix and still use NMS to eliminate low scoring predictions, reducing the quality of their estimated uncertainty for both objects' bounding box and category.

Le \etal~\cite{Uncertainty_Le_2018_ITSC} estimate aleatoric uncertainty in deep object detectors by exploiting \textbf{anchor redundancy}, where multiple per-anchors predictions map to the same object. These predictions are clustered using spatial affinity \textit{before NMS}, and uncertainty measures are estimated using the cluster associated with every output prediction. Finally, a straightforward extension of~\cite{What_Uncertainty_Kendal_2017_NIPS} is typically used to perform \textbf{joint estimation of epistemic and aleatoric} uncertainty in deep object detectors~\cite{Towards_Feng_ITSC_2018, Uncertainty_Kraus_2019_Arxiv}, while still employing NMS to eliminate rather than fuse information from redundant anchors.

Unlike each of the existing methods, BayesOD replaces NMS with Bayesian inference significantly improving the quality of its uncertainty estimates. In addition, BayesOD is the first method to tackle fusion of the category from redundant output anchors, as well as to provide a multivariate extension of Eq.~\eqref{eq:kendal_variance_regression} to estimate the aleatoric uncertainty of objects' bounding boxes.

\section{A Bayesian Formulation For Object Detection:}
\label{sec:proposed_solution}
Throughout this section, the bounding box of an object, represented by its top left and bottom right corners, is denoted as $\BB$, whereas its category, represented by a one-hot vector, is denoted as $\SC$. The index $i$ is used to signify a variable related to the $i$\textsuperscript{th} anchor in the anchor grid. Variables not indexed with $i$ represent inference output clustered over several anchors. Finally, predictions provided by the neural network are denoted with a $\hat{.}$ operator.

\subsection{Computing The Per-Anchor Gaussian Posterior:}
\label{subsec:gaussian}
\noindent\textbf{Computing the uncertainty in the estimated per-anchor bounding box:}
Following \cite{What_Uncertainty_Kendal_2017_NIPS} and using MC-Dropout as a tractable approximation of the integral in Eq.~\eqref{eq:parameter_marginal}, the sufficient statistics of the Gaussian marginal probability distribution describing the estimated per-anchor bounding box $\hat \BB_i \sim \mathcal{N}(\bm\mu(\textbf{x}_i),\Sigma(\textbf{x}_i))$ can be derived as:
\begin{align}
\label{eq:regression_marginal_mean}
&\bm\mu(\textbf{x}_i) = \frac{1}{T} \sum\limits_{t=1}^T f(\textbf{x}_i, \bm\theta_t)\\
\label{eq:regression_marginal_covariance}
&\Sigma_e(\textbf{x}_i) = \frac{1}{T} \left(\sum\limits_{t=1}^T f(\textbf{x}_i, \bm\theta_t)f(\textbf{x}_i, \bm\theta_t)^\intercal\right) - \bm\mu(\textbf{x}_i)\bm\mu(\textbf{x}_i)^\intercal,
\end{align}
where $T$ is the number of times MC-Dropout sampling is performed, and $f(\textbf{x}_i,\bm\theta_t)$ is the bounding box regression output of the neural network for the $t$\textsuperscript{th} MC-Dropout run. The covariance matrix, $\Sigma_e$, captures the epistemic uncertainty in the estimated bounding box $\hat \BB_i$.

Eq.~\eqref{eq:regression_marginal_mean} is sufficient to compute the output mean of the per-anchor bounding box $\hat\BB_i$. However, Eq.~\eqref{eq:regression_marginal_covariance} still needs to account for the aleatoric component of uncertainty, where the final per-anchor output covariance $\Sigma(\textbf{x}_i)$ can be approximated as:
\begin{equation}
\label{eq:covar_update_sum}
\Sigma(\textbf{x}_i) = \Sigma_e(\textbf{x}_i) + \frac{1}{T}\sum\limits_{t=1}^T \Sigma_a(\textbf{x}_i, \bm\theta_t).
\end{equation}
To estimate the full covariance matrix $\Sigma_a(\textbf{x}_i)$, a novel multivariate log likelihood regression loss is derived as:
\begin{multline}
\label{eq:ali_uncertainty}
L_{mv}(\textbf{x}_i, \bm\theta) = \\\frac{1}{2}(f(\textbf{x}_i, \bm\theta) - \textbf{y}_i)^\intercal \Sigma_a(\textbf{x}_i, \bm\theta)^{-1}(f(\textbf{x}_i, \bm\theta) - \textbf{y}_i) \\+  \frac{1}{2}\log\det \Sigma_a(\textbf{x}_i, \bm\theta),
\end{multline}
where $\Sigma_a(\textbf{x}_i, \bm\theta)$ is the predicted per-anchor aleatoric covaraince matrix, $f(\textbf{x}_i, \bm\theta)$ is the predicted per-anchor bounding box, and $\textbf{y}_i$ is the associated regression target. However, the loss in Eq.~\eqref{eq:ali_uncertainty} is found to be numerically unstable. Furthermore, there are no guarantees on the positive definiteness of the predicted covariance matrix $\Sigma_a(\textbf{x}_i, \bm\theta)$. Using the $LDL$ decomposition of $\Sigma_a(\textbf{x}_i, \bm\theta) = L(\textbf{x}_i, \theta)D(\textbf{x}_i, \theta)L(\textbf{x}_i, \theta)^\intercal$, in conjunction with the Cauchy-Schwarz inequality, a numerically stable surrogate loss function is derived as:
\begin{multline}
\label{eq:ali_fixed_uncertainty}
L_{mv}(\textbf{x}_i, \bm\theta) =\\ \frac{1}{2}|| L(\textbf{x}_i, \bm\theta)^{-1}||_F^2||D(\textbf{x}_i, \bm\theta)^{-\frac{1}{2}}( f(\textbf{x}_i, \bm\theta) - \textbf{y}_i)||_2^2\\ +  \frac{1}{2}tr(\log D(\textbf{x}_i, \bm\theta)),
\end{multline}
where $L(\textbf{x}_i, \bm\theta)$ is a lower triangular matrix with ones for its diagonal entries, and $D(\textbf{x}_i, \bm\theta)$ is a diagonal matrix. The loss function in Eq.~\eqref{eq:ali_fixed_uncertainty} is a numerically stable upper bound of the one in Eq.~\eqref{eq:ali_uncertainty} and can guarantee the positive definiteness of $\Sigma_a(\textbf{x}_i, \bm\theta)$ by predicting positive values for the diagonal elements of $D(\textbf{x}_i, \bm\theta)$ through standard activation functions. The final output distributions after incorporating both epistemic and aleatoric covariance estimates are plotted as bounding boxes in the middle image of \fig{fig:bayesian_framework}.\\

\noindent\textbf{Incorporating per-anchor bounding box priors:}
The per-anchor bounding box prior is usually defined based on the training dataset $\mathcal{D}$ as $p(\BB|\textbf{x}_i) \sim \mathcal{N}(\bm\mu_0, \Sigma_0)$. The per-anchor posterior distribution describing the bounding box $\BB$ can then be written as:
\begin{equation}
\label{eq:per_anchor_reg_posterior}
p(\BB | \textbf{x}_i, \mathcal{D}, \hat\BB_i) \propto p(\hat \BB_i| \textbf{x}_i, \mathcal{D}, \BB) p(\BB|\textbf{x}_i, \mathcal{D}). 
\end{equation}
$p(\hat \BB_i| \textbf{x}_i, \mathcal{D}, \BB)$ is a Gaussian likelihood function described by the sufficient statistics $[\bm\mu(\textbf{x}_i), \Sigma(\textbf{x}_i) ]$ in equations Eq.~\eqref{eq:regression_marginal_mean} and Eq.~\eqref{eq:covar_update_sum}. The sufficient statistics can be computed through the multivariate Gaussian conjugate update, as:
\begin{align}
\label{eqn:postCovMC}
\Sigma'(\textbf{x}_i) &= (\Sigma_0^{-1} + \Sigma(\textbf{x}_i)^{-1})^{-1} \\
\label{eqn:postMeanMC}
\bm\mu'(\textbf{x}_i)&= \Sigma'(\textbf{x}_i)(\Sigma_0^{-1} \bm\mu_0 + \Sigma(\textbf{x}_i) \bm\mu(\textbf{x}_i)).
\end{align}

The choice of anchor priors depends on the application, and whether object information is actually available a priori. Since no useful bounding box information is available from our 2D training datasets, a non-informative prior, visually shown in the left image of \fig{fig:bayesian_framework}, is chosen for $\BB$ following ~\cite{gelman2013bayesian}.

\subsection{Computing The Per-Anchor Categorical Posterior:}
\label{subsec:categorical}
\noindent\textbf{Computing the uncertainty in the estimated per-anchor category:}
Since the neural network outputs the parameters of a Categorical distribution rather than one-hot categorical samples, the parameters for the Categorical marginal conditional probability distribution $\hat\SC_i \sim Cat([\hat p_1\dots\hat p_K])$ can be computed as:
\begin{align}
\label{eq:classification_marginal_ps}
\hat p_k = \frac{1}{T} \sum\limits_{t=1}^T \mbox{SoftMax}(g(\textbf{x}_i, \bm\theta_t))_k,
\end{align}
where $\mbox{SoftMax}(.)$ is the soft max function, and $g(\textbf{x}_i, \bm\theta_t)_k$ is the output \textit{logit} of the $k$\textsuperscript{th} category, estimated at the $t$\textsuperscript{th} MC-Dropout run of the neural network. No explicit treatment of the aleatoric classification uncertainty is performed, since it is already contained within the estimated parameters $[\hat p_1\dots\hat p_K]$~\cite{Leveraging_Feng_2018_ITSC}.\\

\noindent\textbf{Incorporating per-anchor category priors:}
For the object category, a Dirichlet distribution is set as a prior over the parameters $\mathcal{P}$ of the categorical distribution $Cat(\mathcal{P})$ generating $\SC$, instead of incorporating a prior distribution directly over the category $\SC$. The posterior distribution of the categorical parameters can be written as:
\begin{equation}
\label{eq:per_anchor_cat_post}
p(\mathcal{P}| \textbf{x}_i, \mathcal{D}, \hat{\bm{Z}}_i) \propto p(\hat{\bm{Z}}_i| \textbf{x}_i, \mathcal{D}, \mathcal{P}) p(\mathcal{P}| \textbf{x}_i, \mathcal{D}),
\end{equation}
where $\mathcal{P}$ is the set of updated parameters $ [p'_1,\dots,p'_K]$, and $\hat{\bm{Z}_i} = [\hat{\bm{  z_1}},\dots,\hat{\bm{z}}_H]$ are $H$ \textbf{i.i.d.} samples from $Cat([\hat p_1,\dots,\hat p_K])$. Since the likelihood function $p(\hat{\bm{Z}_i}| \textbf{x}_i, \mathcal{D}, \mathcal{P})$ is a categorical distribution, the prior distribution $p(\mathcal{P} | \textbf{x}_i, \mathcal{D})$ is chosen to be a Dirichlet distribution allowing a Dirichlet posterior to be computed in closed form as:
\begin{align}
\label{eq:dir_post}
\nonumber p(\mathcal{P}| \textbf{x}_i, \mathcal{D}, \hat{\bm{Z}}_i) &\propto \prod\limits_{k=1}^K p_k^{\alpha_k -1} \prod\limits_{h=1}^H \prod\limits_{k=1}^K p_k^{\hat z_{hk}=1 } \\
& = Dir(\alpha'_1,\ldots,\alpha'_K),
\end{align}
where $\hat z_{hk}$ is the element in instance $\hat{\bm{z}}_h$ corresponding to category $k$, and $[\alpha_k' = \alpha_k + \sum\limits_{h=1}^H \hat z_{hk} \ \forall \ k = 1,\dots, K]$ are the inferred parameters of the Dirichlet posterior distribution. The per-anchor categorical posterior distribution can be written as:
\begin{equation}
\label{eq:cat_post_post}
p(\SC| \textbf{x}_i, \mathcal{D}, \bm Z_i) = Cat([p'_1, \ldots, p'_K]),   
\end{equation}
where $p'_k$  is the mean of the Dirichlet posterior distribution~\cite{gelman2013bayesian} in Eq.~\eqref{eq:dir_post} written as:
\begin{align*}
p'_k = \frac{\alpha'_k}{\sum\limits_{j=1}^K \alpha'_j}.
\end{align*}

Similar to the prior used for the per-anchor bounding box, we choose a non-informative Dirichlet prior for the per-anchor category following~\cite{gelman2013bayesian}. Although non-informative, the prior still serves an essential purpose by allowing the derivation of a Dirichlet posterior in Eq.~\eqref{eq:dir_post}, which will allow the fusion of information from multiple clustered categorical variables in the next section.

\subsection{Bayesian Inference as a Replacement to NMS:} 
Similar to NMS, BayesOD clusters per-anchor outputs from the neural network using spatial affinity. However, all elements in the cluster are then combined regardless of their classification score during inference. Greedy clustering is chosen as it provides adequate performance when compared to standard NMS, while maintaining computational efficiency. For better performing but slower clustering algorithms, see~\cite{Evaluating_Merging_Miller_2018_Arxiv}.

For the remainder of this section, we will continue the derivation for a single anchor cluster containing $M$ anchors. The anchor with the highest categorical score is considered the cluster's center, is indexed by $1$, and is described with the posterior distributions in Eq.~\eqref{eq:per_anchor_reg_posterior} and Eq.~\eqref{eq:per_anchor_cat_post}. The rest of the cluster members are assumed to be measurement outputs from the neural network described by the states $\hat\SC_i$ and $\hat\BB_i$, and are used to update the bounding box and category of the cluster center. Specifically, the final posterior distribution describing an object's bounding box is:
\begin{align}
\label{eq:gaussian_posterior_merging}
\nonumber p(\BB | \mathcal{X}, \mathcal{D}, [\hat\BB_1,\dots,\hat\BB_M])& \propto p(\BB|\textbf{x}_1, \mathcal{D},\hat\BB_1) \prod\limits_{i=2}^M p(\hat{\BB}_i | \textbf{x}_i, \mathcal{D}, \BB)  \\
&= \mathcal{N}(\bm\mu''(\mathcal{X}), \Sigma''(\mathcal{X})),
\end{align}
where $\mathcal{X}$ is the set of inputs $[\textbf{x}_i \ | \ i=1\dots M]$ corresponding to the $M$ cluster members, $p(\BB|\textbf{x}_1, \mathcal{D},\hat\BB_1)$ is the per-anchor posterior distribution of the cluster center, and $p(\BB|\textbf{x}_1, \mathcal{D},\hat\BB_1) \prod\limits_{i=2}^M p(\hat{\BB}_i | \textbf{x}_i, \mathcal{D}, \BB) $ is the likelihood derived through a conditional independence assumption of the $\hat \BB_i$ of the cluster members given $\BB$. The sufficient statistics of Eq.~\eqref{eq:gaussian_posterior_merging} can be estimated in closed form as: 
\begin{align}
\Sigma''(\mathcal{X}) &= \left(\sum\limits_{i=1}^M \Sigma'(\textbf{x}_i)^{-1}\right)^{-1}\\
\bm\mu''(\mathcal{X}) &=  \Sigma''(\mathcal{X}) \left(\sum\limits_{i=1}^M \Sigma'(\textbf{x}_i)^{-1} \bm\mu'(\textbf{x}_i)\right),
\end{align}
where $\bm\mu'(\textbf{x}_i), \Sigma'(\textbf{x}_i)$ are the sufficient statistics of the per anchor posterior distribution derived in Eq.~\eqref{eq:per_anchor_reg_posterior}.

To arrive at the final posterior distribution describing the category $\SC$, a similar analysis can be performed to update the sufficient statistics $\mathcal{P}$ of the cluster center with categorical measurements $[\hat{\bm{Z}}_2,\dots,\hat{\bm{Z}}_m]$ of the rest of the cluster members. Specifically, the posterior probability of $\mathcal{P}$ can be derived as:
\begin{align}
\label{eq:cat_posterior_merging}
\nonumber p(\mathcal{P}| \textbf{x}_i, \mathcal{D}, [\hat{\bm{Z}}_1,\dots,\hat{\bm{Z}}_M]) &\propto p(\mathcal{P}|\textbf{x}_1, \mathcal{D},\hat{\bm{Z}}_1) \prod\limits_{i=2}^M p(\hat{\bm{Z}}_i | \textbf{x}_i, \mathcal{D}, \mathcal{P})\\
&= Dir(\alpha''_1,\dots,\alpha''_K)
\end{align}
 where $\alpha''_{k} = \alpha'_{k} + \sum\limits_{i=2}^M\sum\limits_{h=1}^H \hat z_{ihk} \ \forall \ k=1\dots K$, and the categorical measurements $[\hat{\bm{Z}}_2,\dots,\hat{\bm{Z}}_m]$ are assumed to be \textbf{i.i.d}. In summary, $\alpha''_{k}$ is derived by updating the per-anchor Dirichlet posterior distribution in \eqref{eq:per_anchor_cat_post} of the cluster center with index $i=1$ with categorical measurements $\bm{Z}_2,\dots,\bm{Z}_M$ from all cluster members.
 The final categorical distribution describing the state $\SC$ is then:
\begin{equation}
\label{eq:cat_posterior_final}
p(\SC| \mathcal{X}, \mathcal{D},[\hat{\bm{Z}}_1,\dots,\hat{\bm{Z}}_M] ) = Cat(p''_1,\dots,p''_K),
\end{equation}
where $[p''_1,\dots,p''_K]$ is computed as the mean of the posterior distribution in Eq.~\eqref{eq:cat_posterior_merging}:
\begin{equation}
p''_k = \frac{\alpha''_{k}}{\sum\limits_{j=1}^K \alpha''_{j}},
\end{equation}
Note that every member of the cluster contributes to the estimation of the final bounding box and category states of the object. Furthermore, the output distributions for both the category and bounding box can be updated with object-level priors using the same equations presented in sections \ref{subsec:gaussian} and \ref{subsec:categorical}.  The final output from BayesOD is shown as the rightmost image in \fig{fig:bayesian_framework}.
\begin{table*}[t]
\centering
\resizebox{\textwidth}{!}{
\begin{tabular}{c|c|c|c|c|c|c|c|}
Training Dataset & Testing Dataset & Method & mAP(\%) $\uparrow$ & PDQ Score(\%) $\uparrow$& mGMUE(\%) $\downarrow$ &mCMUE(\%) $\downarrow$\\ \midrule
\multirow{5}{*}{BDD}& \multirow{5}{*}{BDD}& Sampling Free & 36.59 & 33.97 & 44.19 & 28.46 \\
& & Black Box & 36.43& 32.46 & 47.63 &30.45 \\
& & Anchor Redundancy& 32.92 & 29.57 & 48.56 & 35.58 \\
& & Joint Aleatoric-Epistemic& 36.84 & 29.57 & 46.35 & 28.28\\
\rowcolor{Gray}
& & BayesOD& \textbf{38.14} & \textbf{36.79} & \textbf{34.42} & \textbf{24.85}\\ \midrule
\multirow{5}{*}{BDD}& \multirow{5}{*}{Kitti}& Sampling Free & 64.78 & 29.24 & 46.70 & 20.67 \\
& & Black Box & 62.96& 32.26 & 49.23 &22.27 \\
& & Anchor Redundancy& \textbf{64.83} & 29.57 & 48.56 & 35.58 \\
& & Joint Aleatoric-Epistemic& 62.96 & 29.57 & 46.35 & 28.28\\
\rowcolor{Gray}
& & BayesOD& 63.34 & \textbf{35.26} & \textbf{30.06} & \textbf{15.58}\\ \midrule \midrule
\multirow{5}{*}{COCO}& \multirow{5}{*}{COCO}& Sampling Free & 31.89 & 22.43 & 40.39 & 25.76 \\
& & Black Box & 33.71& 21.87 & 45.26 &28.68 \\
& & Anchor Redundancy& 29.94 & 17.63 & 43.74 & 31.13 \\
& & Joint Aleatoric-Epistemic& 32.68 & 23.08 & 42.90 & 26.51\\
\rowcolor{Gray}
& & BayesOD& \textbf{35.41} & \textbf{23.15} & \textbf{30.23} & \textbf{24.13}\\ \midrule
\multirow{5}{*}{COCO}& \multirow{5}{*}{Pascal VOC}& Sampling Free & 54.94 & \textbf{14.18} & 49.49 & 29.63 \\
& & Black Box & 54.67& 12.77 & 48.90 &29.42 \\
& & Anchor Redundancy& 51.56 & 13.06 & 48.67 & 39.64 \\
& & Joint Aleatoric-Epistemic& 55.43& 11.62 & 49.99 & 30.14\\
\rowcolor{Gray}
& & BayesOD& \textbf{56.00} & 13.23 & \textbf{36.36} & \textbf{24.19}\\
\bottomrule   
\end{tabular}
}
\caption{The results of the evaluation of \textit{Sampling Free}~\cite{Leveraging_Feng_2018_ITSC, Uncertainty_Le_2018_ITSC}, \textit{Black Box}~\cite{Dropout_Miller_2018_ICRA, Evaluating_Merging_Miller_2018_Arxiv}, \textit{Anchor Redundancy}~\cite{Uncertainty_Le_2018_ITSC}, and \textit{Joint Aleatoric-Epistemi}c~\cite{Towards_Feng_ITSC_2018, Uncertainty_Kraus_2019_Arxiv} state of the art methods compared to BayesOD.}
\label{table:comparison_studies}
\vspace{-2.5em}
\end{table*}
\section{Experiments and Results}
\label{sec:experiments}
To show the effectiveness of BayesOD in comparison to the state of the art, it is applied to the problem of 2D object detection in image space. The evaluation is based on four commonly used datasets: 
\begin{itemize}
    \item\textbf{Berkley Deep Drive 100K Dataset (BDD)}~\cite{BDD_100K_Yu_2018_Arxiv} road scene dataset, with $80K$ frames used according to the official $70K/10K$ training/validation split. Models trained on BDD are also tested on $7,481$ frames of \textbf{KITTI}~\cite{Kitti_Geiger_2012_CVPR}. Both datasets contain $7$ common road scene object categories.
    \item \textbf{MS COCO}~\cite{COCO_Lin_2014_ECCV} dataset, with $223K$ frames that contain instances from $81$ different object categories, and an official $118K/5K$ training/testing split. Models trained on COCO are also tested on $5,823$ frames from \textbf{Pascal VOC}~\cite{PASCAL_Everingham_2010_IJCV}, which shares $20$ object categories with the COCO dataset.
\end{itemize}
Models used for testing are not allowed to observe instances from the KITTI or Pascal VOC datasets.

All baseline uncertainty estimation methods used in comparison are integrated into the inference process of RetinaNet \cite{Focal_Loss_Lin_2017_ICCV}, trained using the regression loss function in Eq.~\eqref{eq:kendal_variance_regression} to estimate a diagonal bounding box covariance matrix. Full aleatoric covariance matrix results are provided through a second RetinaNet model, trained using the proposed regression loss in Eq.~\eqref{eq:ali_fixed_uncertainty}. For additional information on RetinaNet's training procedure and hyperparamters, see \cite{Focal_Loss_Lin_2017_ICCV}.

\subsection{Evaluation Metrics}
Three evaluation metrics are used to quantify the performance of uncertainty estimation methods in comparison to BayesOD. For performance on the detection task, we use the \textbf{Mean Average Precision (mAP)}~\cite{COCO_Lin_2014_ECCV, PASCAL_Everingham_2010_IJCV,BDD_100K_Yu_2018_Arxiv, Kitti_Geiger_2012_CVPR} at $0.5$ IOU. The maximum mean average precision achievable by a detector is $100\%$. 

The \textbf{Minimum Uncertainty Error (MUE)}~\cite{Evaluating_Merging_Miller_2018_Arxiv} at $0.5$ IOU is used to determine the ability of the detector's estimated uncertainty to discriminate true positives from false positives. The lowest MUE achievable by a detector is $0\%$. We define the Gaussian MUE (GMUE) when the Gaussian entropy  is used, Categorical MUE (CMUE) when the Categorical entropy is used. Finally, we average the GMUE and CMUE over all categories in a testing dataset to arrive to a single value, the Mean (Gaussian or Categorical) MUE (mGMUE or mCMUE).

Finally, we use the newly proposed \textbf{Probability Based Detection Quality (PDQ)}~\cite{Hall_PDQ_2019_ARXIV} to jointly quantify the bounding box and category probability assigned to true positives by the detector. The highest PDQ achievable by a detector is $100\%$, where the PDQ increases as the distributions assigned to a detection better match those of the ground truth instance. For detailed information on the three evaluation metrics, we refer the reader to the~\cite{PASCAL_Everingham_2010_IJCV, Evaluating_Merging_Miller_2018_Arxiv, Hall_PDQ_2019_ARXIV}.

\subsection{Comparison With State of The Art Methods:}
BayesOD is compared against four approaches representing the state of the art methods for uncertainty estimation methods used for object detection. The four approaches are referred to as: \textit{Black Box}~\cite{Dropout_Miller_2018_ICRA, Evaluating_Merging_Miller_2018_Arxiv}, \textit{Sampling Free}~\cite{Uncertainty_Le_2018_ITSC, Leveraging_Feng_2018_ITSC}, \textit{Anchor Redundancy}~\cite{Uncertainty_Le_2018_ITSC}, and \textit{Joint Aleatoric Epistemic}~\cite{Towards_Feng_ITSC_2018}. BayesOD, Black Box, and Joint Aleatoric Epistemic use  $10$ stochastic runs of MC-Dropout, while Sampling Free and Anchor Redundancy use only one non-stochastic run. As such, BayesOD, Black Box, and Joint Aleatoric Epistemic run at a similar frame rate, approximately $4\times$ slower than Sampling Free and Anchor Redundancy. The affinity threshold used for clustering in all methods was set to the $0.5$ IOU, similar to that used for NMS in RetinaNet. The number of categorical samples $H$ in Eq.~\eqref{eq:per_anchor_cat_post} is empirically set to $30$.

Table \ref{table:comparison_studies} shows the results of evaluating the four methods in comparison to BayesOD, on the four testing datasets.
BayesOD is seen to outperform all four methods on mAP when tested on the BDD, COCO and PASCAL VOC datasets by a margin of $0.57 \% - 1.7\% $ over the second best method, but is outperformed on the KITTI dataset by $\sim 1.5\%$ when using the Sampling Free and Anchor Redundancy methods. Such reduction in performance on KITTI is noted with all methods using MC-Dropout, implying that MC-Dropout might hurt mAP performance in cases where the testing dataset is semantically different than the training dataset. 

Similarly, BayesOD also outperforms all four methods on PDQ when tested on the BDD, KITTI and COCO datasets by a margin of $0.07 \% - 3.00\% $ over the second best method. BayesOD is outperformed on the PASCAL VOC dataset by $0.95\%$ when using the sampling free method. Considering the performance only on PDQ, it cannot be determined if a method is assigning lower probability values to false positives.

On the other hand, the mGMUE/mCMUE are capable of providing a quantitative measure of how well the estimated uncertainty can be used to separate correct and incorrect detections \cite{Evaluating_Merging_Miller_2018_Arxiv}. BayesOD provides a significant reduction of $9.77\% - 13.13\%$ in mGMUE over the next best method on all four testing datasets. Combined with BayesOD's performance on the PDQ metric, it can be inferred that BayesOD not only assigns adequate probability to true positives, but also assigns a lower probability to false positives when compared to true positives. Finally, when comparing mCMUE, BayesOD provides a reduction between $1.63\%-5.23\%$ over the next best method on all four datasets.

 
 \begin{table}[!b]
\centering
\resizebox{\columnwidth}{!}{
\begin{tabular}{cccccc}
$\#$& Experiment & mAP(\%) $\uparrow$ & PDQ Score(\%) $\uparrow$ & mGMUE(\%) $\downarrow$ & mCMUE(\%) $\downarrow$ \\ \midrule
1 & Full System &\textbf{35.41} & \textbf{23.15} & 30.23 & \textbf{24.13} \\
2 & Diagonal Covariance& 34.77 & 22.64 & 30.69 & 25.25  \\
3 & Epistemic Only& 34.15 & 22.62 & 35.88 & 26.47 \\
4 & Aleatoric Only&34.12 & 22.67 & \textbf{28.95} & 25.60 \\
5 & Standard NMS&34.70 & 22.65 & 43.19 & 25.10 \\
\bottomrule   
\end{tabular}
}
\caption{The results of ablation studies performed on BayesOD using the COCO dataset for training and testing.}
\label{table:ablation_studies}
\end{table}

\subsection{Ablation Studies:}
Table \ref{table:ablation_studies} shows the results of the mAP, PDQ, mGMUE, and mCMUE for the ablation studies performed on the COCO dataset. The results of the full BayesOD framework can be seen in experiment $\#1$. By analyzing the results of the ablation studies, the following claims are put forth:

\textbf{Learning the off-diagonal elements of the covariance matrix provides slightly better uncertainty estimates for the objects' bounding box.} To support this claim, RetinaNet is trained using the original log likelihood loss in Eq.~\eqref{eq:kendal_variance_regression} instead of the proposed multivariate loss in Eq.~\eqref{eq:ali_fixed_uncertainty}. The results of BayesOD using this original loss formulation are shown in experiment $\#2$. When compared to the full system, an increase of $0.48\%$ is observed in mGMUE. Although the improvement is not substantial, the new proposed loss avoids an explicit independence assumption and allows the neural network to learn to drive the off-diagonal elements of the covariance matrix towards $0$ if needed.

\textbf{Aleatoric uncertainty provides a more discriminative uncertainty estimate for the objects' bounding box over epistemic uncertainty estimated from MC-Dropout.} To support this claim BayesOD is implemented without the update step in Eq.~\eqref{eq:covar_update_sum}, to use only the per-anchor sample variance computed from multiple stochastic runs of MC-Dropout. The results, presented in experiment $\# 3$, show an increase of $5.65\%$ and $2.34\%$ is observed in the mGMUE and mCMUE respectively. Note however that this conclusion is specific to MC-Dropout, and might not be valid for alternative epistemic uncertainty estimation mechanisms.

To provide better insight on the effect of epistemic uncertainty from MC-Dropout on the full system, experiment $\# 4$ is performed by using BayesOD with a single inference run, and without any epistemic uncertainty estimation mechanism. The results show a decrease in mGMUE of $6.93\%$ over experiment $\# 3$, and $1.28\%$ over the full system, further cementing the conclusion that MC-Dropout might not be a good method to estimate epistemic uncertainty in deep object detectors. 

\textbf{Greedy Non-Maximum Suppression is detrimental to the discriminative power of the uncertainty in the objects' bounding box.} To support this claim, the elimination scheme of NMS is selected to retain only cluster centers, while discarding the remaining cluster members. The results presented in experiment $\# 5$ show a large increase of $12.96\%$ mGMUE when compared to the full system. We conclude that merging information from all cluster members into the final object estimate is essential for proper quantification of bounding box uncertainty by a neural network.

\section{Conclusion}
\label{sec:conclusion}
This paper presents BayesOD, a Bayesian approach for estimating the uncertainty in the output of deep object detector. Experiments using BayesOD show that replacing NMS with Bayesian inference and explicitly incorporating full \textit{aleatoric} covariance matrix estimation allows for a much more meaningful estimated category and bounding box uncertainty in deep object detectors. This work aims to pave the path for future research directions that would use BayesOD for active learning, exploration, as well as object tracking. Future work will study the effect of informative priors originating from multiple detectors, temporal information, and different sensors on the perception capabilities of a robotic system.



\bibliographystyle{unsrt}
\bibliography{bayes-od}

\end{document}